\documentclass{article} 
\usepackage{iclr2026_conference,times}

\usepackage{amsmath,amsfonts,bm}

\def\eqref#1{equation~\ref{#1}}

\def\1{\bm{1}}

\DeclareMathAlphabet{\mathsfit}{\encodingdefault}{\sfdefault}{m}{sl}
\SetMathAlphabet{\mathsfit}{bold}{\encodingdefault}{\sfdefault}{bx}{n}

\usepackage{hyperref}
\usepackage{url}

\usepackage{booktabs,siunitx,makecell,array,threeparttable,graphicx}
\usepackage[table]{xcolor} 
\newcolumntype{Z}{S[table-format=3.2,detect-weight,detect-inline-weight]}

\newcommand{\taskrow}[1]{%
  \addlinespace[1pt]%
  \rowcolor{black!6}%
  \multicolumn{9}{l}{\strut\textit{#1}\rule{0pt}{1.1ex}}\\%
  \addlinespace[1pt]%
}

\usepackage{subcaption} 

\newcolumntype{G}{>{\color{gray}}c} 

\usepackage{xspace}

\newcommand{\ours}{\texttt{Reasoning Curriculum}\xspace}
\newcommand{\oursShort}{\texttt{RC}\xspace}

\newcommand{\ourQwenShort}{\texttt{RC-Qwen}\xspace}

\newcolumntype{Z}{S[table-format=2.2,detect-weight,detect-inline-weight]}

\usepackage{enumitem}
\setlist[itemize]{leftmargin=1.2em, labelsep=0.4em, itemsep=0pt, topsep=2pt}

\title{Reasoning Curriculum: Bootstrapping Broad LLM Reasoning from Math}

\author{Bo Pang\textsuperscript{1}, Deqian Kong\textsuperscript{2}, Silvio Savarese\textsuperscript{1}, Caiming Xiong\textsuperscript{1}, Yingbo Zhou\textsuperscript{1} \\
\textsuperscript{1}Salesforce AI Research \quad
\textsuperscript{2}University of California, Los Angeles \\
\texttt{\{b.pang, cxiong, yingbo.zhou\}@salesforce.com}
}
\iclrfinalcopy 
\begin{document}

\maketitle

\begin{abstract}
Reinforcement learning (RL) can elicit strong reasoning in large language models (LLMs), yet most open efforts focus on math and code. We propose \textbf{\ours}, a simple two-stage curriculum that first elicits reasoning skills in pretraining-aligned domains such as math, then adapts and refines these skills across other domains via joint RL. Stage 1 performs a brief cold start and then math-only RL with verifiable rewards to develop reasoning skills. Stage 2 runs joint RL on mixed-domain data to transfer and consolidate these skills. The curriculum is minimal and backbone-agnostic, requiring no specialized reward models beyond standard verifiability checks. Evaluated on Qwen3-4B and Llama-3.1-8B over a multi-domain suite, \ours yields consistent gains. Ablations and a cognitive-skill analysis indicate that both stages are necessary and that math-first elicitation increases cognitive behaviors important for solving complex problems. \ours provides a compact, easy-to-adopt recipe for general reasoning.
\end{abstract}

\section{Introduction}
Recent work has advanced rapidly on eliciting reasoning in large language models (LLMs). Chain-of-Thought (CoT) prompting \citep{wei2022chain} asks models to produce intermediate steps before answering and substantially improves reasoning performance. Building on this idea, proprietary systems train with reinforcement learning (RL) to refine long chains of thought, achieving strong results in competition math and programming \citep{OpenAIo1}. Open-source efforts follow a similar trajectory, reporting competitive performance and exposing training practices to broader scrutiny \citep{qwq2024,guo2025deepseekr1,zeng2025simplerl,deepscaler2025,deepcoder2025}.

Despite this progress, most open-source research concentrates on math and code, domains with abundant data and easily verifiable rewards. General reasoning across diverse domains remains comparatively underexplored. Recent work expands beyond math and code \citep{akter2025nemotroncrossthinkscalingselflearningmath,generalreasoner,cheng2025revisitingreinforcementlearningllm} and focuses on curating data across broad domains, yet effective, cross-domain training strategies for strong reasoning models are still scarce.

We start from a premise suggested by the literature and our preliminary experiments: math is unusually amenable to RL-based skill elicitation. Significant gains can arise even under weak supervision, including spurious or random rewards, and sometimes from very small training sets \citep{spuriousreward,one-shot-rl}. We hypothesize that math serves as an effective driver for discovering core reasoning skills that can later be adapted to other domains through on-policy training.

This paper proposes \textbf{\ours}, a simple two-stage curriculum. Stage 1 elicits reasoning via supervised cold start and math-only RL. Stage 2 transfers and refines the learned skills by running joint RL on a mixed-domain corpus spanning math, STEM, code, simulation, logic, and tabular tasks. The curriculum is intentionally minimal, requires no specialized reward models beyond standard verifiability checks, and applies across backbones.

We evaluate \ours on Qwen3-4B and Llama-3.1-8B. On Qwen, our 4B model consistently outperforms similarly sized baselines and is competitive with, and sometimes exceeds, 32B systems. On Llama, directly porting the Qwen recipe yields only small gains, so we introduce a simple difficulty curriculum within the Math-RL stage (medium then hard). With this change, the curriculum again improves performance across all domains.

We also provide evidence for the mechanism behind \ours through ablations and a cognitive-skill analysis, showing that math-first elicitation increases transferable behaviors and that both stages are necessary for the full gains.

In summary, \ours follows a simple strategy: first develop reasoning skills in pretraining-aligned domains such as math using verifiable rewards, then adapt and refine them across diverse domains with joint RL. This yields a compact, easy to adopt training recipe for general reasoning that consistently improves performance across domains.

\begin{figure}[htbp]
  \centering
  \includegraphics[width=\linewidth]{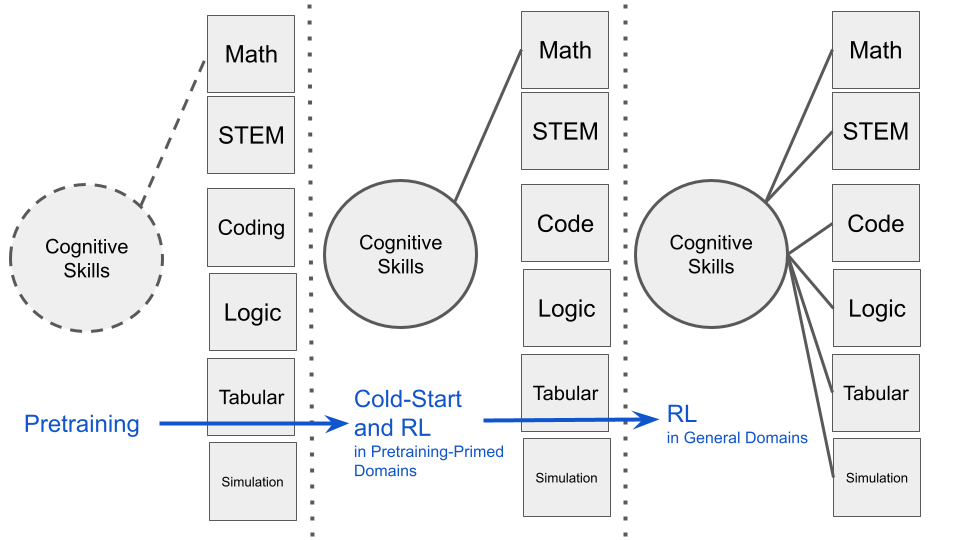}
\caption{Reasoning curriculum overview. Stage 0 (pretraining, not conducted in this work): cognitive skills exist but are weakly expressed on data-rich domains like math. Stage 1 (cold-start + math-only RL): skills are elicited and strengthened in pretraining-primed domains. Stage 2 (joint RL): skills are transferred and refined across general domains (code, logic, tabular, simulation). Blue arrows indicate the training progression.}
  \label{fig:reasoning_curriculum}
\end{figure}

\section{Reasoning Curriculum}

Suppose \((x, y)\) is a question–answer pair and \(z\) is a chain of thought that produces \(y\). The reasoning process often manifests distinct cognitive skills. Four skills, commonly observed in both human solvers and successful LLMs \citep{gandhi2025cognitivebehaviorsenableselfimproving,zeng2025simplerl}, are:
\begin{itemize}
    \item Subgoal setting: decomposing a complex problem into smaller, manageable steps.
    \item Enumeration: considering multiple cases or possibilities.
    \item Backtracking: identifying errors during generation and explicitly revising prior steps.
    \item Verification: checking intermediate results to ensure correctness.
\end{itemize}
While subgoal setting and enumeration frequently appear in most modern LLMs with CoTs, verification and backtracking are often associated with LongCoT models such as Deepseek-R1~\citep{deepseekai2025deepseekr1incentivizingreasoningcapability} and are critical for solving harder problems. Our goal is to increase the use of these skills in general domains and thereby strengthen LLM reasoning.

It is frequently observed that reinforcement learning with verifiable rewards (RLVR) on math data increases the use of these skills and yields substantial gains \citep{zeng2025simplerl,deepscaler2025,deepcoder2025,hu2025open}, even under noisy rewards \citep{shao2025spurious}. Given the readiness of skill elicitation in the math domain, we hypothesize that pretraining already exposes models to these skills in data-rich domains such as math, making them easier to elicit during post-training.

We therefore propose a two-stage reasoning curriculum (Figure~\ref{fig:reasoning_curriculum}). First, we elicit skills on math via a brief cold start followed by reinforcement learning with verifiable rewards. Second, we refine and adapt these skills through joint RL on mixed-domain data to improve general reasoning.

\subsection{Math Training}

\subsubsection{Cold Start}

Given a pretrained LLM, we first perform supervised fine-tuning on a small set of math examples to expose the model to skill-rich thought traces:
\begin{equation}
    \mathcal{J}_{\text{Cold-Start}}(\theta)
    = \mathbb{E}_{(x,z,y)\sim\mathcal{D}_{\text{CS}}}\!\left[\log \pi_{\theta}(y, z \mid x)\right].
\label{eq:cold_start}
\end{equation}

Although recent work explores a zero-RL setup that applies RL without any supervised LongCoT training \citep{OpenReasonerZero2025,zeng2025simplerl}, in practice strong reasoning systems almost always begin with some cold-start supervision. Even within the DeepSeek-R1 line, which popularized the zero-RL idea, widely used variants include supervised components \citep{deepseekai2025deepseekr1incentivizingreasoningcapability}. We therefore adopt a brief cold start. It quickly exposes the model to diverse reasoning skills and creates a realistic setting to study how SFT interacts with RL. Empirically, cold start helps the model imitate multiple cognitive skills, while on-policy RL is still critical to consolidate these behaviors into measurable gains in reasoning performance (see Section~\ref{sec:cognitive_skill} for detailed discussions).

\subsubsection{Math RL}
For RL, Group Relative Policy Optimization (GRPO) \citep{deepseek-math} has become popular due to its efficiency and the success of DeepSeek-R1 \citep{deepseekai2025deepseekr1incentivizingreasoningcapability}. We use the DAPO variant \citep{yu2025dapo}, which introduces several modifications that improve stability and performance:
\begin{equation}
\begin{aligned}
\mathcal{J}_{\text{DAPO}}(\theta) =\quad& \mathbb{E}_{(x,y)\sim \mathcal{D}, \{y_i\}_{i=1}^G\sim \pi_{\theta_\text{old}}(\cdot\mid x)}\\&
\Bigg[\frac{1}{\sum_{i=1}^{G}|y_i|}\sum_{i=1}^{G}\sum_{t=1}^{|y_i|} 
\min \Big( r_{i,t}(\theta) \hat{A}_{i,t},  
\ \text{clip} \Big( r_{i,t}(\theta), 1 - {\varepsilon_{\text{low}}}, 1 + {\varepsilon_{\text{high}}} \Big) \hat{A}_{i,t} \Big) \Bigg]
\\
\text{s.t.}\quad& 0< \Big|\{y_i\mid\texttt{is\_equivalent}(y,y_i)\}\Big|< G,
\label{eq:dapoloss}
\end{aligned}
\end{equation}
where
\begin{equation}
    r_{i,t}(\theta)=\frac{\pi_{\theta}(y_{i,t} \mid x, y_{i,<t})}{\pi_{\theta_{\text{old}}}(y_{i,t} \mid x,y_{i,<t})},\quad\hat{A}_{i,t} = \frac{R_i - \text{mean}(\{R_i\}_{i=1}^G)}{\text{std}(\{R_i\}_{i=1}^G)}.
\label{eq:advantage_calculation}
\end{equation}
The constraint filters groups so that at least one sample is correct and at least one is incorrect, which makes relative advantages meaningful. Also, we omit the KL penalty to encourage exploration.

Following \citet{zeng2025simplerl}, we avoid format rewards that can hinder exploration and use only correctness as the outcome reward:
\begin{equation}
    R(\hat{y}, y) =
    \begin{cases}
        1, & \texttt{is\_equivalent}(\hat{y}, y) \\
        0, & \text{otherwise.}
    \end{cases}
\end{equation}

\subsection{Joint RL}
After the Math-focused stage, we train a single policy with joint RL across our full suite of domains (Math, Code, STEM, Logic, Simulation, Tabular; see Experiments for details). Training uses the same DAPO objective as in Equation~\ref{eq:dapoloss}; only the reward computation differs by domain. Unless noted otherwise, rewards are binary \(R \in \{0,1\}\) (1 if the prediction matches the ground truth, 0 otherwise). Two Logic datasets permit partial credit, so we assign \(R \in (0,1)\) when appropriate (see Experiments~\ref{sec:data}). All rewards are derived automatically from verifiable signals and are therefore low noise, which is the key to stable and effective RL. Following prior work on general reasoning \citep{generalreasoner, cheng2025revisitingreinforcementlearningllm}, we combine three evaluation strategies to accommodate domain-specific answer formats:
\begin{itemize}
    \item Rule-based matching.   Used in Math, Logic, Simulation, and Tabular. The model is prompted to place the final answer in a prescribed format (e.g., \verb|\boxed{}|). We extract and normalize the answer,
    then compare it with the ground-truth for exact or numeric equivalence.
    \item Model-based equivalence.   Used in STEM where questions have free-form answers and deterministic rules are brittle. An LLM is used to compare the model output with the reference answer for semantic equivalence. This method robustly handles phrasing differences while maintaining low reward noise.
    \item Execution-based verification.   Used in Code. The generated function or script is executed against a unit-test suite and receives a reward of 1 only if all tests pass, and 0 otherwise.
\end{itemize}

\section{Experiments}

\subsection{Training Data}
\label{sec:data}

\paragraph{Cold Start Data}
We randomly sample 20k problems from NuminaMath \citep{li2024numinamath} and generate responses with DeepSeek-R1 \citep{deepseekai2025deepseekr1incentivizingreasoningcapability}. We retain 10k examples whose R1 responses produce correct answers and use them for cold-start training.

\paragraph{Reinforcement Learning Data}
Our RL training builds on recent public datasets for LLM reasoning. Early efforts emphasize math \citep{skywork-or1-2025,yu2025dapo,deepscaler2025} and code \citep{deepcoder2025,likaixin2024taco-verified,2025synthetic1,jain2024livecodebench}, while newer releases broaden coverage to STEM, logic, simulation, and tabular reasoning \citep{generalreasoner,akter2025nemotroncrossthinkscalingselflearningmath,logiczebra,li2025codei,cheng2025revisitingreinforcementlearningllm,stojanovski2025reasoninggymreasoningenvironments}. Two resources are especially useful: \citet{cheng2025revisitingreinforcementlearningllm} consolidates multi-domain datasets from prior work, and \citet{stojanovski2025reasoninggymreasoningenvironments} provides a library with 100+ data generators and verifiers. We draw primarily from these public releases and use the standard verifiable rewards they provide. Our training domains are summarized below.

\begin{itemize}
    \item Math. Challenging problems from exams, practice sets, and competitions with verifiable final answers.
    \item STEM. Questions collected from QA sources and refined with LLMs. Subjects span physics, chemistry, business, history and more; answers may be numeric, symbolic expressions, or booleans.
    \item Code. Coding challenges from competitive programming and LeetCode-style datasets with unit tests.
    \item Simulation. Tasks adapted from code-based environments that require procedural simulation within the chain of thought, such as predicting program outputs (forward simulation) or inferring inputs for a given output (backward simulation).
    \item Logic. Datasets emphasizing constraint satisfaction and formal deduction.
    \item Tabular. Problems that require parsing, querying, and reasoning over one or more tables to synthesize the final answer.
\end{itemize}

\subsection{Training Setup}
We experiment with two models: Qwen3-4B \citep{yang2025qwen3technicalreport} and Llama-3.1-8B \citep{grattafiori2024llama3herdmodels} since they strike a practical balance of model performance and training cost.

\textit{Cold-Start SFT.} We use Axolotl \citep{axolotl} with AdamW \citep{loshchilov2017decoupled}. The peak learning rate is \(5\times 10^{-5}\) with 10\% linear warmup, then decays to \(0.1\times\) the peak. Training runs for 4 epochs. The same hyperparameters are used for both backbones.

\textit{Reinforcement Learning.} We use \texttt{verl} \citep{sheng2024hybridflow} with AdamW. The learning rate is \(1\times 10^{-6}\) with 10 warmup steps and then decays to 0. The prompt batch size is 256; for each prompt we sample 16 responses with temperature 1.0. The maximum input length is 4096 tokens and the maximum output length is 8192 tokens.

\subsection{Evaluation Benchmarks}
We evaluate across six domains using widely adopted benchmarks: Math (AIME24; MATH500 \citep{hendrycks2021measuring}), Code (HumanEval; MBPP; LiveCodeBench \citep{chen2021evaluating,austin2021program,jain2024livecodebench}), STEM (GPQA; SuperGPQA \citep{gpqa,supergpqa}), Logic (Zebra; Knights and Knaves; BoxNet \citep{logiczebra,stojanovski2025reasoninggymreasoningenvironments}), Simulation (CodeI/O; CRUXEval \citep{li2025codei,gu2024cruxeval}), and Tabular (HiTab; MultiHiertt; FinQA \citep{cheng2021hitab,zhao2022multihiertt,finqa}).

\subsection{Baselines}  
We compare our models to several recent reasoning models that are trained with public data on math or general domains:
(1) \text{General Reasoner}~\citep{generalreasoner},
(2) \text{SimpleRL-Zoo}~\citep{zeng2025simplerl}
(3) Guru~\citep{cheng2025revisitingreinforcementlearningllm}.
In addition, we also compare reasoning curriculum to two variants where some components are removed: 1) cold start + joint RL where math-RL is removed, 2) direct joint RL where both cold-start and math-RL are removed. These comparisons would help us understand the contributions of each component in our curriculum.

\section{Results}
\subsection{Results on Qwen}

The Qwen results are summarized in Table~\ref{tab:main-baselines-qwen}. Across all domains, the 4B model trained with reasoning curriculum (\ourQwenShort) consistently outperforms similarly sized baselines: Guru-7B, General-Reasoner-7B, and SimpleRL-7B. Despite its smaller size, \ourQwenShort is competitive with, and in several cases exceeds, 32B baselines. Relative to SimpleRL (trained primarily on math), \ourQwenShort matches or surpasses it on math benchmarks and delivers clear gains on most non-math tasks. Compared with Guru-32B (trained on diverse domains and similar data as ours), \ourQwenShort is competitive on the majority of tasks and leads on six benchmarks, supporting our claim that a math-first curriculum followed by joint cross-domain RL yields strong general reasoning in compact models.

\begin{table*}[t]
\centering
\setlength{\tabcolsep}{6pt}
\caption{Evaluation Results on Qwen.}
\label{tab:main-baselines-qwen}
\begin{threeparttable}
\footnotesize
\begin{tabular}{l G G | *{6}{Z}}
\toprule
& \multicolumn{2}{c|}{\textbf{32B}}
& \multicolumn{3}{c}{\textbf{7B}}
& \multicolumn{3}{c}{\textbf{4B}} \\
\cmidrule(lr){2-3}\cmidrule(lr){4-6}\cmidrule(lr){7-9}

\textbf{Task} &
\multicolumn{1}{c}{\makecell{GURU}} &
\multicolumn{1}{c|}{\makecell{SimpleRL}} 
& \multicolumn{1}{c}{\makecell{GURU}} &
\multicolumn{1}{c}{\makecell{General \\ Reasoner}} &
\multicolumn{1}{c}{\makecell{SimpleRL}} &
\multicolumn{1}{c}{\makecell{RL}} &
\multicolumn{1}{c}{\makecell{CS+RL}} &
\multicolumn{1}{c}{\makecell{Reasoning\\ Curriculum}} \\
\midrule

\taskrow{Math}
AIME-24       & 34.89 & 27.20 & 17.50 & 17.08 & 15.60 & 26.56 & 27.71 & \textbf{32.60} \\
Math-500      & 86.00 & 89.60 & 77.25 & 70.40 & 87.00 & 83.20 & 85.20 & \textbf{89.00} \\
\addlinespace[3pt]\cmidrule(lr){1-9}\addlinespace[2pt]

\taskrow{STEM}
GPQA          & 50.63 & 46.46 & 40.78 & 38.64 & 35.98 & 45.83 & 48.99 & \textbf{53.16} \\
SuperGPQA     & 43.60 & 37.73 & 31.80 & 30.64 & 27.29 & 33.00 & 39.60 & \textbf{41.40} \\
\addlinespace[3pt]\cmidrule(lr){1-9}\addlinespace[2pt]

\taskrow{Code}
HumanEval     & 90.85 & 81.25 & 82.62 & 61.12 & 58.08 & 88.79 & 89.55 & \textbf{90.85} \\
LiveCodeBench & 29.30 & 19.80 & 16.49 &  8.51 &  6.72 & 23.66 & 23.21 & \textbf{26.34} \\
MBPP          & 78.80 & 76.75 & 70.00 & 39.80 & 49.60 & 72.40 & 75.80 & \textbf{80.00} \\
\addlinespace[3pt]\cmidrule(lr){1-9}\addlinespace[2pt]

\taskrow{Simulation}
CodeIO        & 12.63 &  9.75 & 15.63 &  7.13 &  6.63 &  6.13 & 14.75 & \textbf{20.63} \\
CruxEval-I    & 80.63 & 72.63 & 61.72 & 63.63 & 56.25 & 70.75 & 78.13 & \textbf{82.13} \\
CruxEval-O    & 88.75 & 67.75 & 71.28 & 56.50 & 58.31 & 71.50 & 76.25 & \textbf{79.75} \\
\addlinespace[3pt]\cmidrule(lr){1-9}\addlinespace[2pt]

\taskrow{Logic}
Knights Knaves
  & 17.62
  & 16.22 
  & 14.43 & 14.73 & 15.26
  & 65.94 & 68.69 & \textbf{71.10} \\
BoxNet
  & 0.12
  & 0.25 
  & 1.06 & 1.60 & 0.78
  & 83.85 & 88.77 & \textbf{93.80} \\
Zebra         & 45.21 &  1.16 & 39.40 &  0.07 &  0.62 & 40.51 & 40.11 & \textbf{44.07} \\
\addlinespace[3pt]\cmidrule(lr){1-9}\addlinespace[2pt]

\taskrow{Tabular}
FinQA         & 46.14 & 45.41 & 34.70 & 34.33 & 35.10 & 42.69 & 44.50 & \textbf{45.14} \\
HiTab         & 82.00 & 69.00 & 74.20 & 54.40 & 50.40 & 73.80 & 71.30 & \textbf{76.60} \\
MultiHiertt   & 55.28 & 52.83 & 44.94 & 31.62 & 37.57 & 52.38 & 50.30 & \textbf{54.02} \\
\bottomrule
\end{tabular}

\begin{tablenotes}[flushleft]\footnotesize
\item RL = direct joint RL; CS+RL = cold-start then joint RL.  
\end{tablenotes}
\end{threeparttable}
\end{table*}


\subsection{Results on Llama}
Table~\ref{tab:main-llama} reports results on Llama. Simply porting the Qwen recipe to Llama-3.1-8B yielded negligible gains, so we introduced two adjustments. First, we initialized from the instruct model (Llama-3.1-8B-Instruct)\footnote{\url{https://huggingface.co/meta-llama/Llama-3.1-8B-Instruct}} rather than the base model, because the base model does not reliably follow instructions, which complicates reward extraction and impedes learning. Second, within the Math-RL stage we added a difficulty curriculum with two sub-stages: medium problems followed by hard problems. This curriculum made learning more stable and enabled a smooth handoff to joint RL. Because most prior work on RL for general reasoning evaluates Qwen models, directly comparable Llama baselines are scarce \citep{generalreasoner,cheng2025revisitingreinforcementlearningllm,OpenReasonerZero2025,akter2025nemotroncrossthinkscalingselflearningmath}. Against our internal baselines, RL (direct joint RL) and CS+RL (cold start + joint RL), the curriculum consistently improves performance across all domains, supporting the claim that math-first elicitation followed by cross-domain RL is effective for Llama.

\begin{table*}[t]
\centering
\setlength{\tabcolsep}{6pt}
\caption{Evaluation Results on Llama.}
\label{tab:main-llama}

\begingroup
\renewcommand{\taskrow}[1]{%
  \addlinespace[1pt]%
  \rowcolor{black!6}%
  \multicolumn{4}{l}{\strut\textit{#1}\rule{0pt}{1.1ex}}\\%
  \addlinespace[1pt]%
}

\begin{threeparttable}
\footnotesize
\begin{tabular}{l *{3}{Z}}
\toprule
\textbf{Task} & \multicolumn{1}{c}{RL} & \multicolumn{1}{c}{CS+RL} & \multicolumn{1}{c}{Reasoning Curriculum} \\
\midrule

\taskrow{Math}
AIME-24       &  7.40 &  9.58 & \textbf{14.37} \\
Math-500      & 55.60 & 69.60 & \textbf{74.40} \\
\addlinespace[3pt]\cmidrule(lr){1-4}\addlinespace[2pt]

\taskrow{STEM}
GPQA          & 32.94 & 35.86 & \textbf{39.90} \\
SuperGPQA     & 27.50 & 29.50 & \textbf{31.70} \\
\addlinespace[3pt]\cmidrule(lr){1-4}\addlinespace[2pt]

\taskrow{Code}
HumanEval     & 70.27 & 69.82 & \textbf{74.24} \\
LiveCodeBench & 15.68 & 17.74 & \textbf{18.46} \\
MBPP          & 60.40 & 58.80 & \textbf{64.00} \\
\addlinespace[3pt]\cmidrule(lr){1-4}\addlinespace[2pt]

\taskrow{Simulation}
CodeIO        & 10.75 & 16.25 & \textbf{17.38} \\
CruxEval-I    & 50.50 & 61.00 & \textbf{65.50} \\
CruxEval-O    & 23.00 & 61.00 & \textbf{60.62} \\
\addlinespace[3pt]\cmidrule(lr){1-4}\addlinespace[2pt]

\taskrow{Logic}
Knights Knaves & 64.47 & 66.67 & \textbf{67.63} \\
BoxNet          & 74.11 & 75.20 & \textbf{96.23} \\
Zebra           & 35.43 & 32.86 & \textbf{41.08} \\
\addlinespace[3pt]\cmidrule(lr){1-4}\addlinespace[2pt]

\taskrow{Tabular}
FinQA        & 27.79 & 33.70 & \textbf{35.33} \\
HiTab        & 74.90 & 75.30 & \textbf{78.30} \\
MultiHiertt  & 40.25 & 38.99 & \textbf{44.05} \\
\bottomrule
\end{tabular}

\begin{tablenotes}[flushleft]\footnotesize
\item RL = direct joint RL; CS+RL = cold-start then joint RL
\end{tablenotes}
\end{threeparttable}
\endgroup
\end{table*}







\begin{figure}[t]
  \centering
  \begin{subfigure}{\linewidth}
    \centering
    \includegraphics[width=\linewidth]{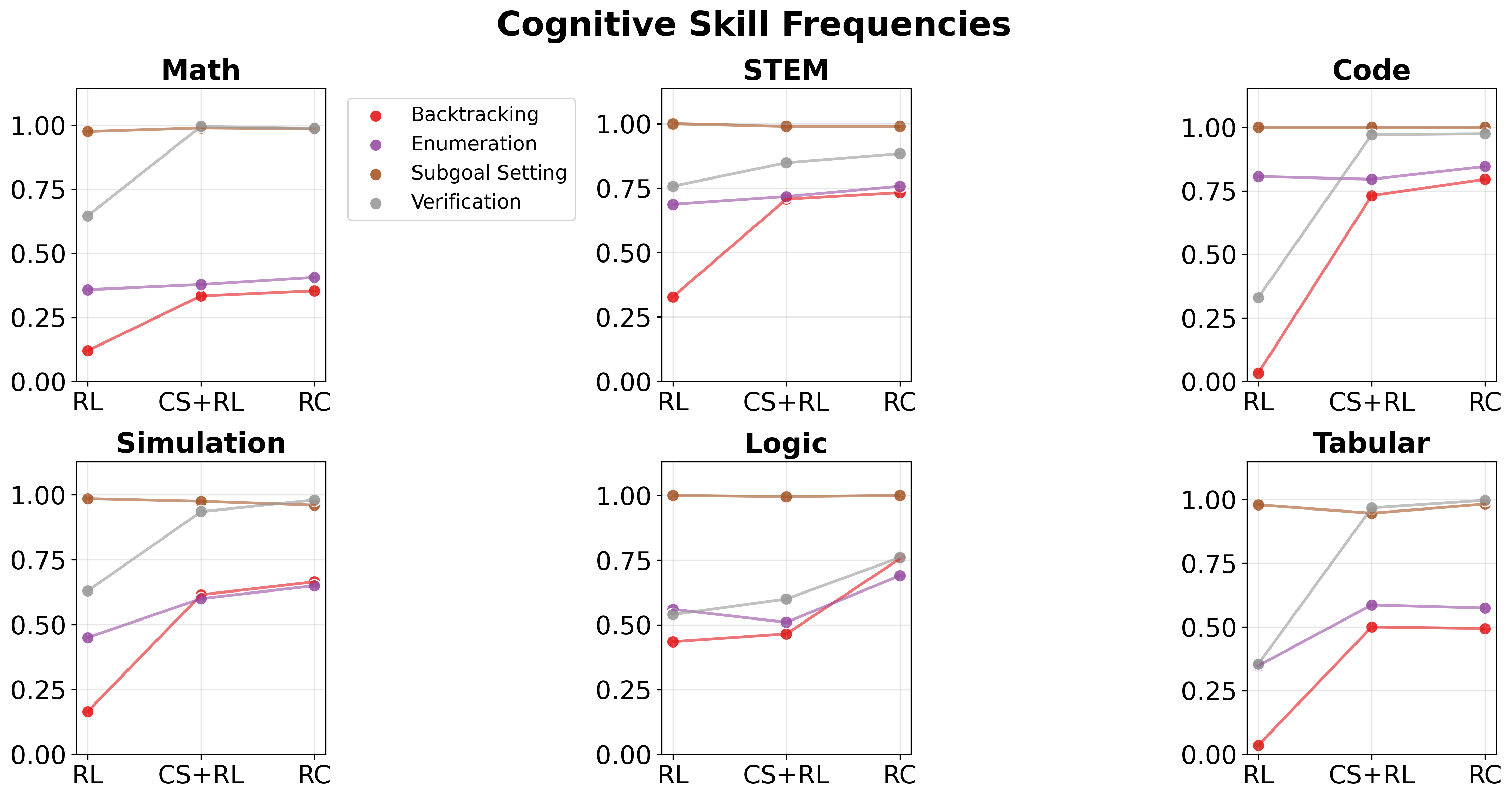}
    \caption{Qwen3-4B results}
    \label{fig:rc-qwen}
  \end{subfigure}

  \vspace{6pt} 

  \begin{subfigure}{\linewidth}
    \centering
    \includegraphics[width=\linewidth]{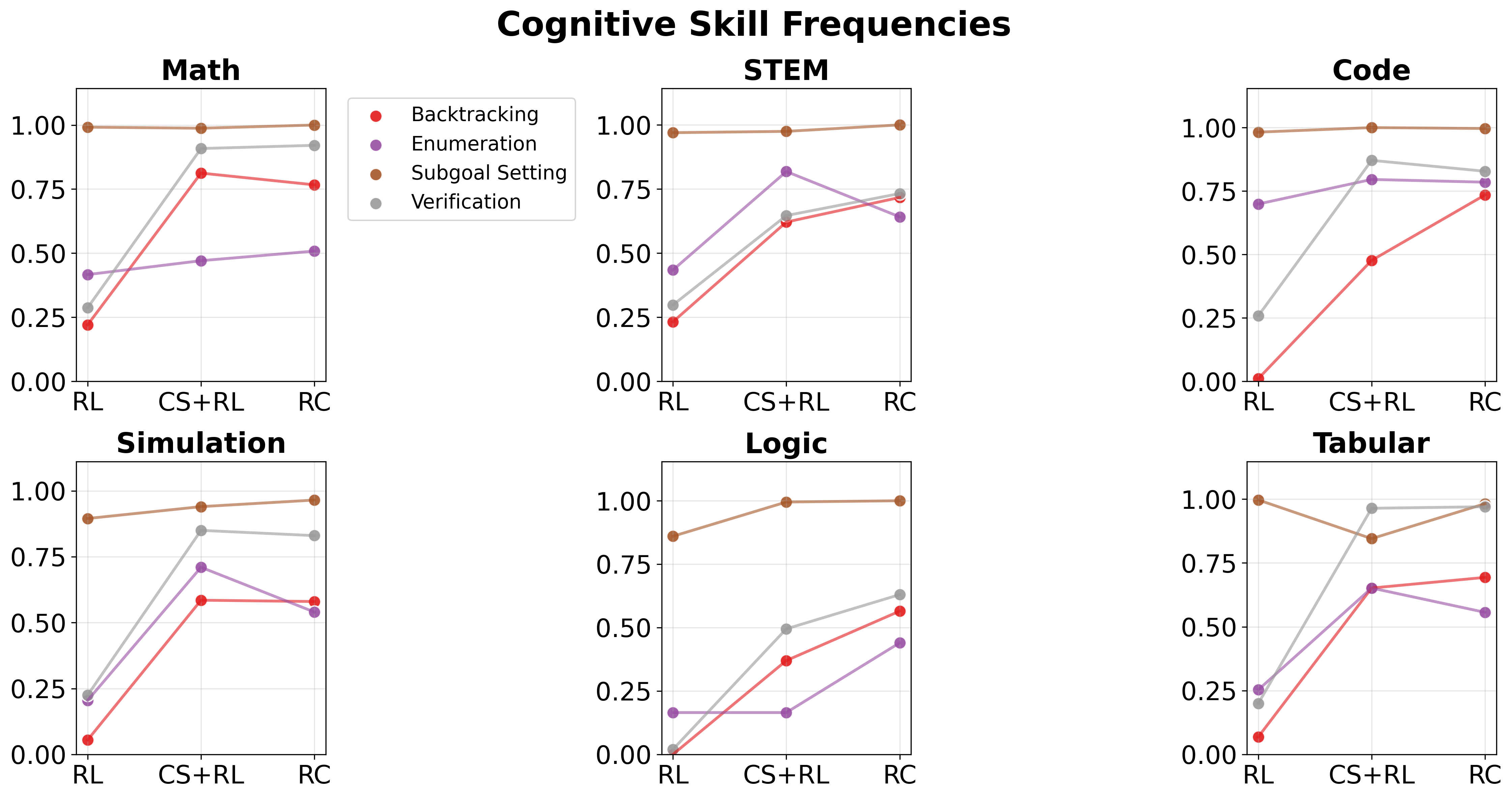}
    \caption{Llama-3.1-8B results}
    \label{fig:rc-llama}
  \end{subfigure}

\caption{Cognitive skill frequencies by training setting. RL = direct joint RL; CS+RL = cold-start then joint RL; RC = reasoning curriculum. Top: Qwen3-4B; bottom: Llama-3.1-8B.}
  \label{fig:skill}
\vspace{-1em}
\end{figure}

\subsection{Cognitive Skills Usage}
\label{sec:cognitive_skill}
We compare cognitive skill frequencies across models trained with Direct Joint RL (RL), Cold-Start + Joint RL (CS+RL), and our Reasoning Curriculum (\oursShort). Following prior work \citep{gandhi2025cognitivebehaviorsenableselfimproving,zeng2025simplerl}, we use GPT-4o-mini to tag four skills: subgoal setting, enumeration, backtracking, and verification. Figure~\ref{fig:skill} summarizes the results (upper: Qwen3-4B; lower: Llama-3.1-8B). Overall, \oursShort increases the frequency of these skills for both backbones, supporting our hypothesis that math-first training improves cognitive skills across domains via the reasoning curriculum. Also, two observations are noteworthy. First, all settings exhibit a similarly high rate of subgoal setting (often near \(100\%\)), which suggests that it is necessary but not sufficient for solving complex problems. Second, CS+RL can show comparable rates of advanced skills in certain domains (for example, backtracking in Tabular for Qwen and verification and backtracking in Simulation for Llama). This suggests that Cold-Start helps models quickly imitate surface-level reasoning patterns, but on-policy training in the Math-RL stage appears important for fully consolidating the skills and converting them into the performance gains observed under the full \oursShort pipeline.

\subsection{Ablations}
We ablate the components of the reasoning curriculum. Table~\ref{tab:ablation} reports average performance. Removing the Math-RL stage, that is, using CS+RL (Cold-Start followed by Joint RL), reduces performance relative to the full curriculum. Removing Cold-Start as well, i.e., direct joint RL, leads to a further drop. The same pattern is observed for both Qwen and Llama models. These results indicate that each component contributes meaningfully to the performance of reasoning curriculum.

\begin{table*}[t]
\centering
\setlength{\tabcolsep}{6pt}
\caption{Ablations on training curriculum.}
\label{tab:ablation}
\begin{threeparttable}
\footnotesize
\begin{tabular}{l *{2}{Z}}
\toprule
\textbf{Ablation} & \multicolumn{1}{c}{\textbf{Qwen3-4B}} & \multicolumn{1}{c}{\textbf{Llama-3.1-8B}} \\
\midrule
Reasoning Curriculum   & {\bfseries 61.29} & {\bfseries 51.45} \\
$-\,$Math-RL           & 57.68             & 46.99 \\
$-\,$Math-RL,\ $-\,$CS & 55.06             & 41.94 \\
\bottomrule
\end{tabular}
\begin{tablenotes}[flushleft]\footnotesize
\item $-\,$Math-RL removes math RL; $-\,$CS further removes cold-start.
\end{tablenotes}
\end{threeparttable}
\end{table*}


\begin{figure}[t]
  \centering
  \begin{subfigure}{0.92\linewidth}
    \centering
    \includegraphics[width=\linewidth]{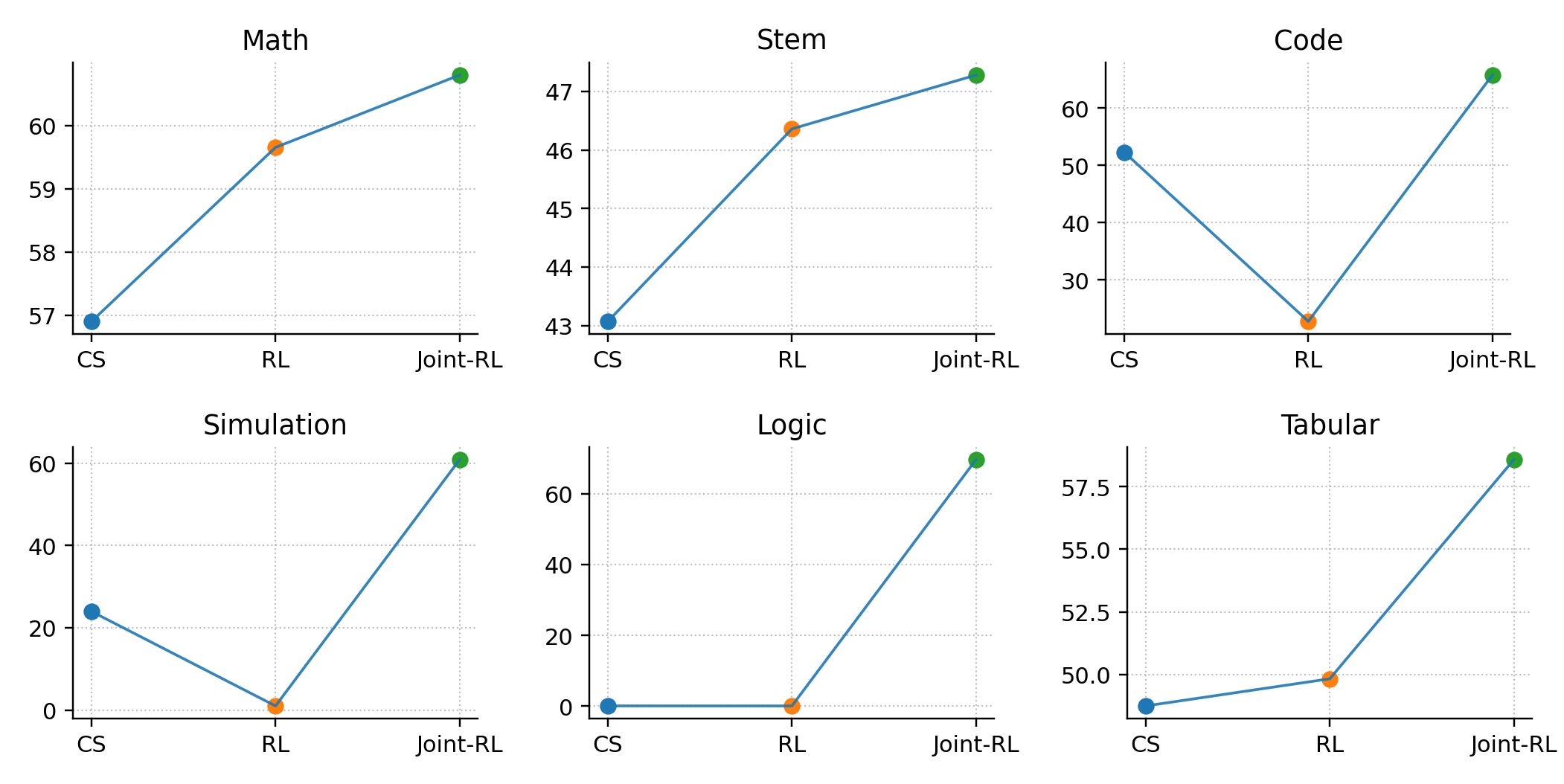}
    \caption{Qwen3-4B results}
    \label{fig:rc-qwen}
  \end{subfigure}

  \vspace{6pt} 

  \begin{subfigure}{0.92\linewidth}
    \centering
    \includegraphics[width=\linewidth]{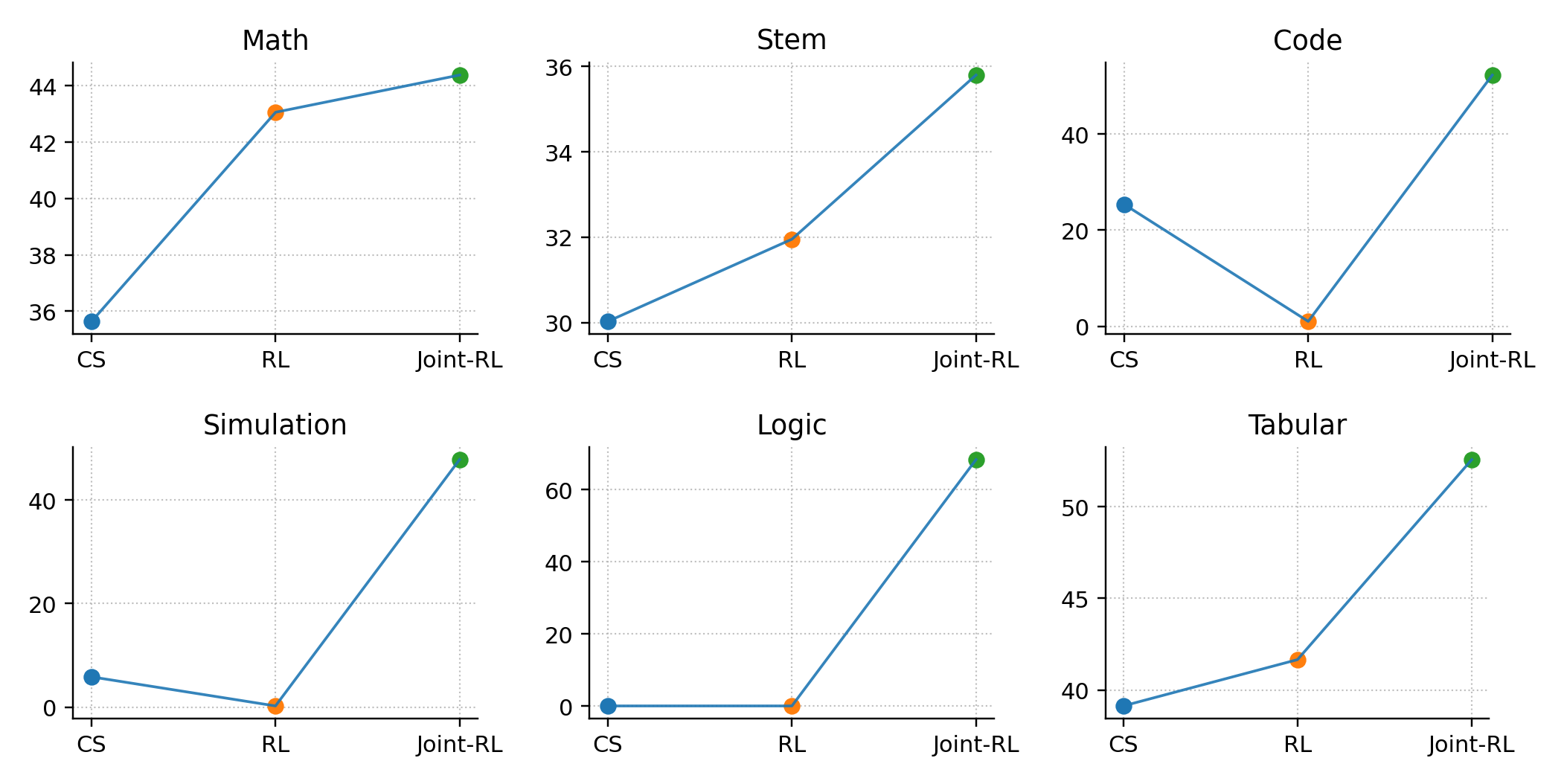}
    \caption{Llama-3.1-8B results}
    \label{fig:rc-llama}
  \end{subfigure}
\caption{Trends across curriculum stages by task. CS = Cold-Start; RL = Math-RL; Joint-RL = RL on mixed-domain data. Top: Qwen3-4B; bottom: Llama-3.1-8B. Each point shows the average score within a domain at each stage.}
  \label{fig:trend}
\vspace{-1em}
\end{figure}

\subsection{Improvements across Reasoning Curriculum}
\label{sec:rc-trend}
We track performance across the curriculum stages (Cold-Start, Math-RL, and Joint-RL) in Figure~\ref{fig:trend} (top: Qwen3-4B; bottom: Llama-3.1-8B). In each sub-figure, the \(y\)-axis is the average score within a domain and the \(x\)-axis indexes the curriculum stage. Three patterns are consistent across both backbones.
First, in Math, STEM, and Tabular, scores improve stage by stage: Math-RL exceeds Cold-Start, and Joint-RL further improves over Math-RL, suggesting shared reasoning representations across these domains.
Second, in Simulation and Code, Math-RL reduces performance relative to Cold-Start even though both stages use only math data, indicating possible overfitting to math. Joint-RL however recovers the drop, and the full curriculum still outperforms the variant that skips Math-RL (see the CS+RL columns in Tables~\ref{tab:main-baselines-qwen} and \ref{tab:main-llama}).
Third, in Logic, performance is near zero after Cold-Start and Math-RL, implying that logic requires domain-specific training. Nevertheless, these stages appear to have a latent positive effect: under the full curriculum, logic accuracy surpasses direct joint RL (compare the RL column with Reasoning Curriculum in Tables~\ref{tab:main-baselines-qwen} and \ref{tab:main-llama}).


\section{Related Work}

\subsection{LLM Reasoning}
A key breakthrough in eliciting reasoning from LLMs is Chain-of-Thought (CoT) prompting \citep{wei2022chain}, which asks models to produce intermediate steps before the final answer. Building on this foundation, recent proprietary models have pushed the boundaries of LLM reasoning by combining massive model scale with large-scale RL. OpenAI's GPT-o1~\citep{OpenAIo1}, for instance, leverages RL to explore and refine long, complex reasoning chains. This approach has demonstrated unprecedented performance on highly challenging domains like competitive math and programming.

The success of this paradigm has inspired the open-source efforts to develop similar capabilities. Models like QwQ~\citep{qwq2024} and DeepSeek-R1~\citep{guo2025deepseekr1} take a similar RL approach and achieve results competitive with leading proprietary models. These efforts have also helped demystify the training process. Community ablations scrutinize when zero or minimal warm-up succeeds and how base model choice affects outcomes \citep{zeng2025simplerl}. There is also evidence that careful scaling and length control can push small models to strong results, for example DeepScaleR-1.5B and DeepCoder-14B, which report competitive performance on verifiable benchmarks \citep{deepscaler2025,deepcoder2025}. Intriguingly, recent studies show that substantial gains on math can be triggered by weak or even misleading reward signals, including rewards that are random or known to be incorrect~\citep{spuriousreward}, and in extreme cases by training on a single example~\citep{one-shot-rl}. This sensitivity of math reasoning to RL supervision motivates our approach: we leverage these dynamics to improve reasoning across domains through a cross-domain reasoning curriculum.


\subsection{Reasoning Across Domains}
Despite rapid progress, most open research concentrates on math and code, where a large amount of data is available and rewards are easily verifiable. Recent efforts have begun to expand coverage beyond these areas. \cite{akter2025nemotroncrossthinkscalingselflearningmath} and \cite{generalreasoner} curate STEM datasets with verifiable rewards, exploiting the ease of multiple-choice verification and using LLMs to normalize and compare answers across varied surface forms. Building on such resources, \citet{cheng2025revisitingreinforcementlearningllm} introduce Guru, which further incorporates logic, simulation, and tabular domains. Collectively, these works advance data collection, cleaning, and cross-domain evaluation, revealing distinct performance patterns across tasks. In our work, we leverage these multi-domain resources and other logic datasets to study how to train a strong reasoning model across domains.


\section{Conclusion}
We introduced \ours, a minimal two-stage curriculum that first elicits reasoning skills in math through cold start and RL, then adapts and refines them with joint RL across diverse domains. On Qwen3-4B and Llama-3.1-8B, \ours delivers consistent multi-domain gains. 
Ablations show that both stages are necessary, and a cognitive-skill analysis indicates increased use of advanced behaviors such as verification and backtracking. The recipe is backbone-agnostic and relies only on standard verifiability checks, which makes it easy to adopt.

\bibliography{iclr2026_conference.bib}

\begin{thebibliography}{41}
\providecommand{\natexlab}[1]{#1}
\providecommand{\url}[1]{\texttt{#1}}
\expandafter\ifx\csname urlstyle\endcsname\relax
  \providecommand{\doi}[1]{doi: #1}\else
  \providecommand{\doi}{doi: \begingroup \urlstyle{rm}\Url}\fi

\bibitem[Akter et~al.(2025)Akter, Prabhumoye, Novikov, Han, Lin, Bakhturina, Nyberg, Choi, Patwary, Shoeybi, and Catanzaro]{akter2025nemotroncrossthinkscalingselflearningmath}
Syeda~Nahida Akter, Shrimai Prabhumoye, Matvei Novikov, Seungju Han, Ying Lin, Evelina Bakhturina, Eric Nyberg, Yejin Choi, Mostofa Patwary, Mohammad Shoeybi, and Bryan Catanzaro.
\newblock Nemotron-crossthink: Scaling self-learning beyond math reasoning, 2025.
\newblock URL \url{https://arxiv.org/abs/2504.13941}.

\bibitem[Austin et~al.(2021)Austin, Odena, Nye, Bosma, Michalewski, Dohan, Jiang, Cai, Terry, Le, et~al.]{austin2021program}
Jacob Austin, Augustus Odena, Maxwell Nye, Maarten Bosma, Henryk Michalewski, David Dohan, Ellen Jiang, Carrie Cai, Michael Terry, Quoc Le, et~al.
\newblock Program synthesis with large language models.
\newblock \emph{arXiv preprint arXiv:2108.07732}, 2021.

\bibitem[Axolotl(2025)]{axolotl}
Axolotl.
\newblock Axolotl: Open source fine-tuning, 2025.
\newblock URL \url{https://github.com/axolotl-ai-cloud/axolotl}.
\newblock Accessed: 2025-01-30.

\bibitem[Chen et~al.(2021)Chen, Tworek, Jun, Yuan, Pinto, Kaplan, Edwards, Burda, Joseph, Brockman, et~al.]{chen2021evaluating}
Mark Chen, Jerry Tworek, Heewoo Jun, Qiming Yuan, Henrique Ponde De~Oliveira Pinto, Jared Kaplan, Harri Edwards, Yuri Burda, Nicholas Joseph, Greg Brockman, et~al.
\newblock Evaluating large language models trained on code.
\newblock \emph{arXiv preprint arXiv:2107.03374}, 2021.

\bibitem[Chen et~al.(2022)Chen, Chen, Smiley, Shah, Borova, Langdon, Moussa, Beane, Huang, Routledge, and Wang]{finqa}
Zhiyu Chen, Wenhu Chen, Charese Smiley, Sameena Shah, Iana Borova, Dylan Langdon, Reema Moussa, Matt Beane, Ting-Hao Huang, Bryan Routledge, and William~Yang Wang.
\newblock Finqa: A dataset of numerical reasoning over financial data, 2022.
\newblock URL \url{https://arxiv.org/abs/2109.00122}.

\bibitem[Cheng et~al.(2021)Cheng, Dong, Wang, Jia, Guo, Gao, Han, Lou, and Zhang]{cheng2021hitab}
Zhoujun Cheng, Haoyu Dong, Zhiruo Wang, Ran Jia, Jiaqi Guo, Yan Gao, Shi Han, Jian-Guang Lou, and Dongmei Zhang.
\newblock Hitab: A hierarchical table dataset for question answering and natural language generation.
\newblock \emph{arXiv preprint arXiv:2108.06712}, 2021.

\bibitem[Cheng et~al.(2025)Cheng, Hao, Liu, Zhou, Xie, Yao, Bian, Zhuang, Dey, Zha, Gu, Zhou, Wang, Li, Fan, She, Gao, Saparov, Li, Killian, Yurochkin, Liu, Xing, and Hu]{cheng2025revisitingreinforcementlearningllm}
Zhoujun Cheng, Shibo Hao, Tianyang Liu, Fan Zhou, Yutao Xie, Feng Yao, Yuexin Bian, Yonghao Zhuang, Nilabjo Dey, Yuheng Zha, Yi~Gu, Kun Zhou, Yuqi Wang, Yuan Li, Richard Fan, Jianshu She, Chengqian Gao, Abulhair Saparov, Haonan Li, Taylor~W. Killian, Mikhail Yurochkin, Zhengzhong Liu, Eric~P. Xing, and Zhiting Hu.
\newblock Revisiting reinforcement learning for llm reasoning from a cross-domain perspective, 2025.
\newblock URL \url{https://arxiv.org/abs/2506.14965}.

\bibitem[DeepSeek-AI(2025)]{deepseekai2025deepseekr1incentivizingreasoningcapability}
DeepSeek-AI.
\newblock Deepseek-r1: Incentivizing reasoning capability in llms via reinforcement learning, 2025.
\newblock URL \url{https://arxiv.org/abs/2501.12948}.

\bibitem[Gandhi et~al.(2025)Gandhi, Chakravarthy, Singh, Lile, and Goodman]{gandhi2025cognitivebehaviorsenableselfimproving}
Kanishk Gandhi, Ayush Chakravarthy, Anikait Singh, Nathan Lile, and Noah~D. Goodman.
\newblock Cognitive behaviors that enable self-improving reasoners, or, four habits of highly effective stars, 2025.
\newblock URL \url{https://arxiv.org/abs/2503.01307}.

\bibitem[Grattafiori et~al.(2024)Grattafiori, Dubey, Jauhri, Pandey, Kadian, Al-Dahle, Letman, Mathur, Schelten, Vaughan, Yang, Fan, Goyal, Hartshorn, Yang, Mitra, Sravankumar, Korenev, Hinsvark, Rao, Zhang, Rodriguez, Gregerson, Spataru, Roziere, Biron, Tang, Chern, Caucheteux, Nayak, Bi, Marra, McConnell, Keller, Touret, Wu, Wong, Ferrer, Nikolaidis, Allonsius, Song, Pintz, Livshits, Wyatt, Esiobu, Choudhary, Mahajan, Garcia-Olano, Perino, Hupkes, Lakomkin, AlBadawy, Lobanova, Dinan, Smith, Radenovic, Guzmán, Zhang, Synnaeve, Lee, Anderson, Thattai, Nail, Mialon, Pang, Cucurell, Nguyen, Korevaar, Xu, Touvron, Zarov, Ibarra, Kloumann, Misra, Evtimov, Zhang, Copet, Lee, Geffert, Vranes, Park, Mahadeokar, Shah, van~der Linde, Billock, Hong, Lee, Fu, Chi, Huang, Liu, Wang, Yu, Bitton, Spisak, Park, Rocca, Johnstun, Saxe, Jia, Alwala, Prasad, Upasani, Plawiak, Li, Heafield, Stone, El-Arini, Iyer, Malik, Chiu, Bhalla, Lakhotia, Rantala-Yeary, van~der Maaten, Chen, Tan, Jenkins, Martin, Madaan, Malo, Blecher,
  Landzaat, de~Oliveira, Muzzi, Pasupuleti, Singh, Paluri, Kardas, Tsimpoukelli, Oldham, Rita, Pavlova, Kambadur, Lewis, Si, Singh, Hassan, Goyal, Torabi, Bashlykov, Bogoychev, Chatterji, Zhang, Duchenne, Çelebi, Alrassy, Zhang, Li, Vasic, Weng, Bhargava, Dubal, Krishnan, Koura, Xu, He, Dong, Srinivasan, Ganapathy, Calderer, Cabral, Stojnic, Raileanu, Maheswari, Girdhar, Patel, Sauvestre, Polidoro, Sumbaly, Taylor, Silva, Hou, Wang, Hosseini, Chennabasappa, Singh, Bell, Kim, Edunov, Nie, Narang, Raparthy, Shen, Wan, Bhosale, Zhang, Vandenhende, Batra, Whitman, Sootla, Collot, Gururangan, Borodinsky, Herman, Fowler, Sheasha, Georgiou, Scialom, Speckbacher, Mihaylov, Xiao, Karn, Goswami, Gupta, Ramanathan, Kerkez, Gonguet, Do, Vogeti, Albiero, Petrovic, Chu, Xiong, Fu, Meers, Martinet, Wang, Wang, Tan, Xia, Xie, Jia, Wang, Goldschlag, Gaur, Babaei, Wen, Song, Zhang, Li, Mao, Coudert, Yan, Chen, Papakipos, Singh, Srivastava, Jain, Kelsey, Shajnfeld, Gangidi, Victoria, Goldstand, Menon, Sharma, Boesenberg,
  Baevski, Feinstein, Kallet, Sangani, Teo, Yunus, Lupu, Alvarado, Caples, Gu, Ho, Poulton, Ryan, Ramchandani, Dong, Franco, Goyal, Saraf, Chowdhury, Gabriel, Bharambe, Eisenman, Yazdan, James, Maurer, Leonhardi, Huang, Loyd, Paola, Paranjape, Liu, Wu, Ni, Hancock, Wasti, Spence, Stojkovic, Gamido, Montalvo, Parker, Burton, Mejia, Liu, Wang, Kim, Zhou, Hu, Chu, Cai, Tindal, Feichtenhofer, Gao, Civin, Beaty, Kreymer, Li, Adkins, Xu, Testuggine, David, Parikh, Liskovich, Foss, Wang, Le, Holland, Dowling, Jamil, Montgomery, Presani, Hahn, Wood, Le, Brinkman, Arcaute, Dunbar, Smothers, Sun, Kreuk, Tian, Kokkinos, Ozgenel, Caggioni, Kanayet, Seide, Florez, Schwarz, Badeer, Swee, Halpern, Herman, Sizov, Guangyi, Zhang, Lakshminarayanan, Inan, Shojanazeri, Zou, Wang, Zha, Habeeb, Rudolph, Suk, Aspegren, Goldman, Zhan, Damlaj, Molybog, Tufanov, Leontiadis, Veliche, Gat, Weissman, Geboski, Kohli, Lam, Asher, Gaya, Marcus, Tang, Chan, Zhen, Reizenstein, Teboul, Zhong, Jin, Yang, Cummings, Carvill, Shepard, McPhie,
  Torres, Ginsburg, Wang, Wu, U, Saxena, Khandelwal, Zand, Matosich, Veeraraghavan, Michelena, Li, Jagadeesh, Huang, Chawla, Huang, Chen, Garg, A, Silva, Bell, Zhang, Guo, Yu, Moshkovich, Wehrstedt, Khabsa, Avalani, Bhatt, Mankus, Hasson, Lennie, Reso, Groshev, Naumov, Lathi, Keneally, Liu, Seltzer, Valko, Restrepo, Patel, Vyatskov, Samvelyan, Clark, Macey, Wang, Hermoso, Metanat, Rastegari, Bansal, Santhanam, Parks, White, Bawa, Singhal, Egebo, Usunier, Mehta, Laptev, Dong, Cheng, Chernoguz, Hart, Salpekar, Kalinli, Kent, Parekh, Saab, Balaji, Rittner, Bontrager, Roux, Dollar, Zvyagina, Ratanchandani, Yuvraj, Liang, Alao, Rodriguez, Ayub, Murthy, Nayani, Mitra, Parthasarathy, Li, Hogan, Battey, Wang, Howes, Rinott, Mehta, Siby, Bondu, Datta, Chugh, Hunt, Dhillon, Sidorov, Pan, Mahajan, Verma, Yamamoto, Ramaswamy, Lindsay, Lindsay, Feng, Lin, Zha, Patil, Shankar, Zhang, Zhang, Wang, Agarwal, Sajuyigbe, Chintala, Max, Chen, Kehoe, Satterfield, Govindaprasad, Gupta, Deng, Cho, Virk, Subramanian, Choudhury,
  Goldman, Remez, Glaser, Best, Koehler, Robinson, Li, Zhang, Matthews, Chou, Shaked, Vontimitta, Ajayi, Montanez, Mohan, Kumar, Mangla, Ionescu, Poenaru, Mihailescu, Ivanov, Li, Wang, Jiang, Bouaziz, Constable, Tang, Wu, Wang, Wu, Gao, Kleinman, Chen, Hu, Jia, Qi, Li, Zhang, Zhang, Adi, Nam, Yu, Wang, Zhao, Hao, Qian, Li, He, Rait, DeVito, Rosnbrick, Wen, Yang, Zhao, and Ma]{grattafiori2024llama3herdmodels}
Aaron Grattafiori, Abhimanyu Dubey, Abhinav Jauhri, Abhinav Pandey, Abhishek Kadian, Ahmad Al-Dahle, Aiesha Letman, Akhil Mathur, Alan Schelten, Alex Vaughan, Amy Yang, Angela Fan, Anirudh Goyal, Anthony Hartshorn, Aobo Yang, Archi Mitra, Archie Sravankumar, Artem Korenev, Arthur Hinsvark, Arun Rao, Aston Zhang, Aurelien Rodriguez, Austen Gregerson, Ava Spataru, Baptiste Roziere, Bethany Biron, Binh Tang, Bobbie Chern, Charlotte Caucheteux, Chaya Nayak, Chloe Bi, Chris Marra, Chris McConnell, Christian Keller, Christophe Touret, Chunyang Wu, Corinne Wong, Cristian~Canton Ferrer, Cyrus Nikolaidis, Damien Allonsius, Daniel Song, Danielle Pintz, Danny Livshits, Danny Wyatt, David Esiobu, Dhruv Choudhary, Dhruv Mahajan, Diego Garcia-Olano, Diego Perino, Dieuwke Hupkes, Egor Lakomkin, Ehab AlBadawy, Elina Lobanova, Emily Dinan, Eric~Michael Smith, Filip Radenovic, Francisco Guzmán, Frank Zhang, Gabriel Synnaeve, Gabrielle Lee, Georgia~Lewis Anderson, Govind Thattai, Graeme Nail, Gregoire Mialon, Guan Pang,
  Guillem Cucurell, Hailey Nguyen, Hannah Korevaar, Hu~Xu, Hugo Touvron, Iliyan Zarov, Imanol~Arrieta Ibarra, Isabel Kloumann, Ishan Misra, Ivan Evtimov, Jack Zhang, Jade Copet, Jaewon Lee, Jan Geffert, Jana Vranes, Jason Park, Jay Mahadeokar, Jeet Shah, Jelmer van~der Linde, Jennifer Billock, Jenny Hong, Jenya Lee, Jeremy Fu, Jianfeng Chi, Jianyu Huang, Jiawen Liu, Jie Wang, Jiecao Yu, Joanna Bitton, Joe Spisak, Jongsoo Park, Joseph Rocca, Joshua Johnstun, Joshua Saxe, Junteng Jia, Kalyan~Vasuden Alwala, Karthik Prasad, Kartikeya Upasani, Kate Plawiak, Ke~Li, Kenneth Heafield, Kevin Stone, Khalid El-Arini, Krithika Iyer, Kshitiz Malik, Kuenley Chiu, Kunal Bhalla, Kushal Lakhotia, Lauren Rantala-Yeary, Laurens van~der Maaten, Lawrence Chen, Liang Tan, Liz Jenkins, Louis Martin, Lovish Madaan, Lubo Malo, Lukas Blecher, Lukas Landzaat, Luke de~Oliveira, Madeline Muzzi, Mahesh Pasupuleti, Mannat Singh, Manohar Paluri, Marcin Kardas, Maria Tsimpoukelli, Mathew Oldham, Mathieu Rita, Maya Pavlova, Melanie Kambadur,
  Mike Lewis, Min Si, Mitesh~Kumar Singh, Mona Hassan, Naman Goyal, Narjes Torabi, Nikolay Bashlykov, Nikolay Bogoychev, Niladri Chatterji, Ning Zhang, Olivier Duchenne, Onur Çelebi, Patrick Alrassy, Pengchuan Zhang, Pengwei Li, Petar Vasic, Peter Weng, Prajjwal Bhargava, Pratik Dubal, Praveen Krishnan, Punit~Singh Koura, Puxin Xu, Qing He, Qingxiao Dong, Ragavan Srinivasan, Raj Ganapathy, Ramon Calderer, Ricardo~Silveira Cabral, Robert Stojnic, Roberta Raileanu, Rohan Maheswari, Rohit Girdhar, Rohit Patel, Romain Sauvestre, Ronnie Polidoro, Roshan Sumbaly, Ross Taylor, Ruan Silva, Rui Hou, Rui Wang, Saghar Hosseini, Sahana Chennabasappa, Sanjay Singh, Sean Bell, Seohyun~Sonia Kim, Sergey Edunov, Shaoliang Nie, Sharan Narang, Sharath Raparthy, Sheng Shen, Shengye Wan, Shruti Bhosale, Shun Zhang, Simon Vandenhende, Soumya Batra, Spencer Whitman, Sten Sootla, Stephane Collot, Suchin Gururangan, Sydney Borodinsky, Tamar Herman, Tara Fowler, Tarek Sheasha, Thomas Georgiou, Thomas Scialom, Tobias Speckbacher,
  Todor Mihaylov, Tong Xiao, Ujjwal Karn, Vedanuj Goswami, Vibhor Gupta, Vignesh Ramanathan, Viktor Kerkez, Vincent Gonguet, Virginie Do, Vish Vogeti, Vítor Albiero, Vladan Petrovic, Weiwei Chu, Wenhan Xiong, Wenyin Fu, Whitney Meers, Xavier Martinet, Xiaodong Wang, Xiaofang Wang, Xiaoqing~Ellen Tan, Xide Xia, Xinfeng Xie, Xuchao Jia, Xuewei Wang, Yaelle Goldschlag, Yashesh Gaur, Yasmine Babaei, Yi~Wen, Yiwen Song, Yuchen Zhang, Yue Li, Yuning Mao, Zacharie~Delpierre Coudert, Zheng Yan, Zhengxing Chen, Zoe Papakipos, Aaditya Singh, Aayushi Srivastava, Abha Jain, Adam Kelsey, Adam Shajnfeld, Adithya Gangidi, Adolfo Victoria, Ahuva Goldstand, Ajay Menon, Ajay Sharma, Alex Boesenberg, Alexei Baevski, Allie Feinstein, Amanda Kallet, Amit Sangani, Amos Teo, Anam Yunus, Andrei Lupu, Andres Alvarado, Andrew Caples, Andrew Gu, Andrew Ho, Andrew Poulton, Andrew Ryan, Ankit Ramchandani, Annie Dong, Annie Franco, Anuj Goyal, Aparajita Saraf, Arkabandhu Chowdhury, Ashley Gabriel, Ashwin Bharambe, Assaf Eisenman, Azadeh
  Yazdan, Beau James, Ben Maurer, Benjamin Leonhardi, Bernie Huang, Beth Loyd, Beto~De Paola, Bhargavi Paranjape, Bing Liu, Bo~Wu, Boyu Ni, Braden Hancock, Bram Wasti, Brandon Spence, Brani Stojkovic, Brian Gamido, Britt Montalvo, Carl Parker, Carly Burton, Catalina Mejia, Ce~Liu, Changhan Wang, Changkyu Kim, Chao Zhou, Chester Hu, Ching-Hsiang Chu, Chris Cai, Chris Tindal, Christoph Feichtenhofer, Cynthia Gao, Damon Civin, Dana Beaty, Daniel Kreymer, Daniel Li, David Adkins, David Xu, Davide Testuggine, Delia David, Devi Parikh, Diana Liskovich, Didem Foss, Dingkang Wang, Duc Le, Dustin Holland, Edward Dowling, Eissa Jamil, Elaine Montgomery, Eleonora Presani, Emily Hahn, Emily Wood, Eric-Tuan Le, Erik Brinkman, Esteban Arcaute, Evan Dunbar, Evan Smothers, Fei Sun, Felix Kreuk, Feng Tian, Filippos Kokkinos, Firat Ozgenel, Francesco Caggioni, Frank Kanayet, Frank Seide, Gabriela~Medina Florez, Gabriella Schwarz, Gada Badeer, Georgia Swee, Gil Halpern, Grant Herman, Grigory Sizov, Guangyi, Zhang, Guna
  Lakshminarayanan, Hakan Inan, Hamid Shojanazeri, Han Zou, Hannah Wang, Hanwen Zha, Haroun Habeeb, Harrison Rudolph, Helen Suk, Henry Aspegren, Hunter Goldman, Hongyuan Zhan, Ibrahim Damlaj, Igor Molybog, Igor Tufanov, Ilias Leontiadis, Irina-Elena Veliche, Itai Gat, Jake Weissman, James Geboski, James Kohli, Janice Lam, Japhet Asher, Jean-Baptiste Gaya, Jeff Marcus, Jeff Tang, Jennifer Chan, Jenny Zhen, Jeremy Reizenstein, Jeremy Teboul, Jessica Zhong, Jian Jin, Jingyi Yang, Joe Cummings, Jon Carvill, Jon Shepard, Jonathan McPhie, Jonathan Torres, Josh Ginsburg, Junjie Wang, Kai Wu, Kam~Hou U, Karan Saxena, Kartikay Khandelwal, Katayoun Zand, Kathy Matosich, Kaushik Veeraraghavan, Kelly Michelena, Keqian Li, Kiran Jagadeesh, Kun Huang, Kunal Chawla, Kyle Huang, Lailin Chen, Lakshya Garg, Lavender A, Leandro Silva, Lee Bell, Lei Zhang, Liangpeng Guo, Licheng Yu, Liron Moshkovich, Luca Wehrstedt, Madian Khabsa, Manav Avalani, Manish Bhatt, Martynas Mankus, Matan Hasson, Matthew Lennie, Matthias Reso, Maxim
  Groshev, Maxim Naumov, Maya Lathi, Meghan Keneally, Miao Liu, Michael~L. Seltzer, Michal Valko, Michelle Restrepo, Mihir Patel, Mik Vyatskov, Mikayel Samvelyan, Mike Clark, Mike Macey, Mike Wang, Miquel~Jubert Hermoso, Mo~Metanat, Mohammad Rastegari, Munish Bansal, Nandhini Santhanam, Natascha Parks, Natasha White, Navyata Bawa, Nayan Singhal, Nick Egebo, Nicolas Usunier, Nikhil Mehta, Nikolay~Pavlovich Laptev, Ning Dong, Norman Cheng, Oleg Chernoguz, Olivia Hart, Omkar Salpekar, Ozlem Kalinli, Parkin Kent, Parth Parekh, Paul Saab, Pavan Balaji, Pedro Rittner, Philip Bontrager, Pierre Roux, Piotr Dollar, Polina Zvyagina, Prashant Ratanchandani, Pritish Yuvraj, Qian Liang, Rachad Alao, Rachel Rodriguez, Rafi Ayub, Raghotham Murthy, Raghu Nayani, Rahul Mitra, Rangaprabhu Parthasarathy, Raymond Li, Rebekkah Hogan, Robin Battey, Rocky Wang, Russ Howes, Ruty Rinott, Sachin Mehta, Sachin Siby, Sai~Jayesh Bondu, Samyak Datta, Sara Chugh, Sara Hunt, Sargun Dhillon, Sasha Sidorov, Satadru Pan, Saurabh Mahajan,
  Saurabh Verma, Seiji Yamamoto, Sharadh Ramaswamy, Shaun Lindsay, Shaun Lindsay, Sheng Feng, Shenghao Lin, Shengxin~Cindy Zha, Shishir Patil, Shiva Shankar, Shuqiang Zhang, Shuqiang Zhang, Sinong Wang, Sneha Agarwal, Soji Sajuyigbe, Soumith Chintala, Stephanie Max, Stephen Chen, Steve Kehoe, Steve Satterfield, Sudarshan Govindaprasad, Sumit Gupta, Summer Deng, Sungmin Cho, Sunny Virk, Suraj Subramanian, Sy~Choudhury, Sydney Goldman, Tal Remez, Tamar Glaser, Tamara Best, Thilo Koehler, Thomas Robinson, Tianhe Li, Tianjun Zhang, Tim Matthews, Timothy Chou, Tzook Shaked, Varun Vontimitta, Victoria Ajayi, Victoria Montanez, Vijai Mohan, Vinay~Satish Kumar, Vishal Mangla, Vlad Ionescu, Vlad Poenaru, Vlad~Tiberiu Mihailescu, Vladimir Ivanov, Wei Li, Wenchen Wang, Wenwen Jiang, Wes Bouaziz, Will Constable, Xiaocheng Tang, Xiaojian Wu, Xiaolan Wang, Xilun Wu, Xinbo Gao, Yaniv Kleinman, Yanjun Chen, Ye~Hu, Ye~Jia, Ye~Qi, Yenda Li, Yilin Zhang, Ying Zhang, Yossi Adi, Youngjin Nam, Yu, Wang, Yu~Zhao, Yuchen Hao, Yundi
  Qian, Yunlu Li, Yuzi He, Zach Rait, Zachary DeVito, Zef Rosnbrick, Zhaoduo Wen, Zhenyu Yang, Zhiwei Zhao, and Zhiyu Ma.
\newblock The llama 3 herd of models, 2024.
\newblock URL \url{https://arxiv.org/abs/2407.21783}.

\bibitem[Gu et~al.(2024)Gu, Rozi{\`e}re, Leather, Solar-Lezama, Synnaeve, and Wang]{gu2024cruxeval}
Alex Gu, Baptiste Rozi{\`e}re, Hugh Leather, Armando Solar-Lezama, Gabriel Synnaeve, and Sida~I Wang.
\newblock Cruxeval: A benchmark for code reasoning, understanding and execution.
\newblock \emph{arXiv preprint arXiv:2401.03065}, 2024.

\bibitem[Guo et~al.(2025)Guo, Yang, Zhang, Song, Zhang, Xu, Zhu, Ma, Wang, Bi, et~al.]{guo2025deepseekr1}
Daya Guo, Dejian Yang, Haowei Zhang, Junxiao Song, Ruoyu Zhang, Runxin Xu, Qihao Zhu, Shirong Ma, Peiyi Wang, Xiao Bi, et~al.
\newblock Deepseek-r1: Incentivizing reasoning capability in llms via reinforcement learning.
\newblock \emph{arXiv preprint arXiv:2501.12948}, 2025.

\bibitem[He et~al.(2025)He, Liu, Liu, Yan, Wang, Cheng, Zhang, Zhang, Xu, Shen, Li, Zeng, Wei, Cheng, An, Liu, and Zhou]{skywork-or1-2025}
Jujie He, Jiacai Liu, Chris~Yuhao Liu, Rui Yan, Chaojie Wang, Peng Cheng, Xiaoyu Zhang, Fuxiang Zhang, Jiacheng Xu, Wei Shen, Siyuan Li, Liang Zeng, Tianwen Wei, Cheng Cheng, Bo~An, Yang Liu, and Yahui Zhou.
\newblock Skywork open reasoner series.
\newblock \url{https://capricious-hydrogen-41c.notion.site/Skywork-Open-Reaonser-Series-1d0bc9ae823a80459b46c149e4f51680}, 2025.
\newblock Notion Blog.

\bibitem[Hendrycks et~al.(2021)Hendrycks, Burns, Kadavath, Arora, Basart, Tang, Song, and Steinhardt]{hendrycks2021measuring}
Dan Hendrycks, Collin Burns, Saurav Kadavath, Akul Arora, Steven Basart, Eric Tang, Dawn Song, and Jacob Steinhardt.
\newblock Measuring mathematical problem solving with the math dataset.
\newblock \emph{arXiv preprint arXiv:2103.03874}, 2021.

\bibitem[Hu et~al.(2025{\natexlab{a}})Hu, Zhang, Han, Jiang, and Xiangyu~Zhang]{OpenReasonerZero2025}
Jingcheng Hu, Yinmin Zhang, Qi~Han, Daxin Jiang, and Heung-Yeung~Shum Xiangyu~Zhang.
\newblock Open-reasoner-zero: An open source approach to scaling reinforcement learning on the base model.
\newblock \url{https://github.com/Open-Reasoner-Zero/Open-Reasoner-Zero}, 2025{\natexlab{a}}.

\bibitem[Hu et~al.(2025{\natexlab{b}})Hu, Zhang, Han, Jiang, Zhang, and Shum]{hu2025open}
Jingcheng Hu, Yinmin Zhang, Qi~Han, Daxin Jiang, Xiangyu Zhang, and Heung-Yeung Shum.
\newblock Open-reasoner-zero: An open source approach to scaling up reinforcement learning on the base model.
\newblock \emph{arXiv preprint arXiv:2503.24290}, 2025{\natexlab{b}}.

\bibitem[Jain et~al.(2024)Jain, Han, Gu, Li, Yan, Zhang, Wang, Solar-Lezama, Sen, and Stoica]{jain2024livecodebench}
Naman Jain, King Han, Alex Gu, Wen-Ding Li, Fanjia Yan, Tianjun Zhang, Sida Wang, Armando Solar-Lezama, Koushik Sen, and Ion Stoica.
\newblock Livecodebench: Holistic and contamination free evaluation of large language models for code.
\newblock \emph{arXiv preprint arXiv:2403.07974}, 2024.

\bibitem[Li et~al.(2024)Li, Beeching, Tunstall, Lipkin, Soletskyi, Huang, Rasul, Yu, Jiang, Shen, et~al.]{li2024numinamath}
Jia Li, Edward Beeching, Lewis Tunstall, Ben Lipkin, Roman Soletskyi, Shengyi Huang, Kashif Rasul, Longhui Yu, Albert~Q Jiang, Ziju Shen, et~al.
\newblock Numinamath: The largest public dataset in ai4maths with 860k pairs of competition math problems and solutions.
\newblock \emph{Hugging Face repository}, 13:\penalty0 9, 2024.

\bibitem[Li et~al.(2025)Li, Guo, Yang, Xu, Wu, and He]{li2025codei}
Junlong Li, Daya Guo, Dejian Yang, Runxin Xu, Yu~Wu, and Junxian He.
\newblock Codei/o: Condensing reasoning patterns via code input-output prediction.
\newblock \emph{arXiv preprint arXiv:2502.07316}, 2025.

\bibitem[Li(2024)]{likaixin2024taco-verified}
Kaixin Li.
\newblock Verified taco problems.
\newblock \url{https://huggingface.co/datasets/likaixin/TACO-verified}, 2024.
\newblock URL \url{https://huggingface.co/datasets/likaixin/TACO-verified}.

\bibitem[Lin et~al.(2025)Lin, Bras, Richardson, Sabharwal, Poovendran, Clark, and Choi]{logiczebra}
Bill~Yuchen Lin, Ronan~Le Bras, Kyle Richardson, Ashish Sabharwal, Radha Poovendran, Peter Clark, and Yejin Choi.
\newblock Zebralogic: On the scaling limits of llms for logical reasoning, 2025.
\newblock URL \url{https://arxiv.org/abs/2502.01100}.

\bibitem[Loshchilov \& Hutter(2017)Loshchilov and Hutter]{loshchilov2017decoupled}
Ilya Loshchilov and Frank Hutter.
\newblock Decoupled weight decay regularization.
\newblock \emph{arXiv preprint arXiv:1711.05101}, 2017.

\bibitem[Luo et~al.(2025{\natexlab{a}})Luo, Tan, Huang, Patel, Ariyak, Wu, Shi, Xin, Cai, Weber, Zhang, Li, Popa, and Stoica]{deepcoder2025}
Michael Luo, Sijun Tan, Roy Huang, Ameen Patel, Alpay Ariyak, Qingyang Wu, Xiaoxiang Shi, Rachel Xin, Colin Cai, Maurice Weber, Ce~Zhang, Li~Erran Li, Raluca~Ada Popa, and Ion Stoica.
\newblock Deepcoder: A fully open-source 14b coder at o3-mini level, 2025{\natexlab{a}}.
\newblock Notion Blog.

\bibitem[Luo et~al.(2025{\natexlab{b}})Luo, Tan, Wong, Shi, Tang, Roongta, Cai, Luo, Li, Popa, and Stoica]{deepscaler2025}
Michael Luo, Sijun Tan, Justin Wong, Xiaoxiang Shi, William~Y. Tang, Manan Roongta, Colin Cai, Jeffrey Luo, Li~Erran Li, Raluca~Ada Popa, and Ion Stoica.
\newblock Deepscaler: Surpassing o1-preview with a 1.5b model by scaling rl, 2025{\natexlab{b}}.
\newblock Notion Blog.

\bibitem[Ma et~al.(2025)Ma, Liu, Jiang, Zhang, Ma, and Chen]{generalreasoner}
Xueguang Ma, Qian Liu, Dongfu Jiang, Ge~Zhang, Zejun Ma, and Wenhu Chen.
\newblock General-reasoner: Advancing llm reasoning across all domains.
\newblock \url{https://github.com/TIGER-AI-Lab/General-Reasoner/blob/main/General_Reasoner.pdf}, 2025.

\bibitem[Mattern et~al.(2025)Mattern, Jaghouar, Basra, Straube, Ferrante, Gabriel, Ong, Weisser, and Hagemann]{2025synthetic1}
Justus Mattern, Sami Jaghouar, Manveer Basra, Jannik Straube, Matthew~Di Ferrante, Felix Gabriel, Jack~Min Ong, Vincent Weisser, and Johannes Hagemann.
\newblock Synthetic-1: Two million collaboratively generated reasoning traces from deepseek-r1, 2025.
\newblock URL \url{https://www.primeintellect.ai/blog/synthetic-1-release}.

\bibitem[OpenAI(2024)]{OpenAIo1}
OpenAI.
\newblock {OpenAI o1 System Card}.
\newblock \url{https://openai.com/index/openai-o1-system-card/}, 2024.

\bibitem[Rein et~al.(2023)Rein, Hou, Stickland, Petty, Pang, Dirani, Michael, and Bowman]{gpqa}
David Rein, Betty~Li Hou, Asa~Cooper Stickland, Jackson Petty, Richard~Yuanzhe Pang, Julien Dirani, Julian Michael, and Samuel~R. Bowman.
\newblock Gpqa: A graduate-level google-proof q\&a benchmark, 2023.
\newblock URL \url{https://arxiv.org/abs/2311.12022}.

\bibitem[Shao et~al.(2025{\natexlab{a}})Shao, Li, Xin, Geng, Wang, Oh, Du, Lambert, Min, Krishna, Tsvetkov, Hajishirzi, Koh, and Zettlemoyer]{shao2025spurious}
Rulin Shao, Shuyue~Stella Li, Rui Xin, Scott Geng, Yiping Wang, Sewoong Oh, Simon~Shaolei Du, Nathan Lambert, Sewon Min, Ranjay Krishna, Yulia Tsvetkov, Hannaneh Hajishirzi, Pang~Wei Koh, and Luke Zettlemoyer.
\newblock Spurious rewards: Rethinking training signals in rlvr, 2025{\natexlab{a}}.
\newblock Notion Blog.

\bibitem[Shao et~al.(2025{\natexlab{b}})Shao, Li, Xin, Geng, Wang, Oh, Du, Lambert, Min, Krishna, Tsvetkov, Hajishirzi, Koh, and Zettlemoyer]{spuriousreward}
Rulin Shao, Shuyue~Stella Li, Rui Xin, Scott Geng, Yiping Wang, Sewoong Oh, Simon~Shaolei Du, Nathan Lambert, Sewon Min, Ranjay Krishna, Yulia Tsvetkov, Hannaneh Hajishirzi, Pang~Wei Koh, and Luke Zettlemoyer.
\newblock Spurious rewards: Rethinking training signals in rlvr, 2025{\natexlab{b}}.
\newblock Notion Blog.

\bibitem[Shao et~al.(2024)Shao, Wang, Zhu, Xu, Song, Zhang, Li, Wu, and Guo]{deepseek-math}
Zhihong Shao, Peiyi Wang, Qihao Zhu, Runxin Xu, Junxiao Song, Mingchuan Zhang, Y.K. Li, Y.~Wu, and Daya Guo.
\newblock Deepseekmath: Pushing the limits of mathematical reasoning in open language models, 2024.
\newblock URL \url{https://arxiv.org/abs/2402.03300}.

\bibitem[Sheng et~al.(2024)Sheng, Zhang, Ye, Wu, Zhang, Zhang, Peng, Lin, and Wu]{sheng2024hybridflow}
Guangming Sheng, Chi Zhang, Zilingfeng Ye, Xibin Wu, Wang Zhang, Ru~Zhang, Yanghua Peng, Haibin Lin, and Chuan Wu.
\newblock Hybridflow: A flexible and efficient rlhf framework.
\newblock \emph{arXiv preprint arXiv:2409.19256}, 2024.

\bibitem[Stojanovski et~al.(2025)Stojanovski, Stanley, Sharratt, Jones, Adefioye, Kaddour, and Köpf]{stojanovski2025reasoninggymreasoningenvironments}
Zafir Stojanovski, Oliver Stanley, Joe Sharratt, Richard Jones, Abdulhakeem Adefioye, Jean Kaddour, and Andreas Köpf.
\newblock Reasoning gym: Reasoning environments for reinforcement learning with verifiable rewards, 2025.
\newblock URL \url{https://arxiv.org/abs/2505.24760}.

\bibitem[Team et~al.(2025)Team, Du, Yao, Ma, Wang, Zheng, Zhu, Liu, Liang, Jin, Wei, Zheng, Deng, Gavin, Jia, Jiang, Liao, Li, Li, Li, Li, Li, Ma, Ni, Que, Wang, Wen, Wu, Hsing, Xu, Yang, Wang, Zhou, Bai, Bu, Cai, Chen, Chen, Cheng, Cheng, Ding, Huang, Huang, Li, Li, Li, Liang, Lin, Lin, Ma, Pang, Peng, Peng, Qi, Qiu, Qu, Quan, Tan, Wang, Wang, Wang, Wang, Wang, Xu, Yang, Yuan, Yue, Zhan, Zhang, Zhang, Zhang, Zhang, Zhang, Zhao, Zheng, Zhong, Gao, Li, Liu, Liu, Liu, Ni, Peng, Qin, Su, Wang, Wang, Yang, Yang, Cao, Yue, Zhang, Zhou, Liu, Lin, Huang, and Zhang]{supergpqa}
P~Team, Xinrun Du, Yifan Yao, Kaijing Ma, Bingli Wang, Tianyu Zheng, King Zhu, Minghao Liu, Yiming Liang, Xiaolong Jin, Zhenlin Wei, Chujie Zheng, Kaixin Deng, Shawn Gavin, Shian Jia, Sichao Jiang, Yiyan Liao, Rui Li, Qinrui Li, Sirun Li, Yizhi Li, Yunwen Li, David Ma, Yuansheng Ni, Haoran Que, Qiyao Wang, Zhoufutu Wen, Siwei Wu, Tyshawn Hsing, Ming Xu, Zhenzhu Yang, Zekun~Moore Wang, Junting Zhou, Yuelin Bai, Xingyuan Bu, Chenglin Cai, Liang Chen, Yifan Chen, Chengtuo Cheng, Tianhao Cheng, Keyi Ding, Siming Huang, Yun Huang, Yaoru Li, Yizhe Li, Zhaoqun Li, Tianhao Liang, Chengdong Lin, Hongquan Lin, Yinghao Ma, Tianyang Pang, Zhongyuan Peng, Zifan Peng, Qige Qi, Shi Qiu, Xingwei Qu, Shanghaoran Quan, Yizhou Tan, Zili Wang, Chenqing Wang, Hao Wang, Yiya Wang, Yubo Wang, Jiajun Xu, Kexin Yang, Ruibin Yuan, Yuanhao Yue, Tianyang Zhan, Chun Zhang, Jinyang Zhang, Xiyue Zhang, Xingjian Zhang, Yue Zhang, Yongchi Zhao, Xiangyu Zheng, Chenghua Zhong, Yang Gao, Zhoujun Li, Dayiheng Liu, Qian Liu, Tianyu Liu, Shiwen
  Ni, Junran Peng, Yujia Qin, Wenbo Su, Guoyin Wang, Shi Wang, Jian Yang, Min Yang, Meng Cao, Xiang Yue, Zhaoxiang Zhang, Wangchunshu Zhou, Jiaheng Liu, Qunshu Lin, Wenhao Huang, and Ge~Zhang.
\newblock Supergpqa: Scaling llm evaluation across 285 graduate disciplines, 2025.
\newblock URL \url{https://arxiv.org/abs/2502.14739}.

\bibitem[Team(2024)]{qwq2024}
Qwen Team.
\newblock Qwq: Reflect deeply on the boundaries of the unknown, 2024.
\newblock URL \url{https://qwenlm.github.io/blog/qwq-32b-preview/}.

\bibitem[Wang et~al.(2025)Wang, Yang, Zeng, Ren, Liu, Peng, Cheng, He, Wang, Gao, Chen, Wang, Du, and Shen]{one-shot-rl}
Yiping Wang, Qing Yang, Zhiyuan Zeng, Liliang Ren, Liyuan Liu, Baolin Peng, Hao Cheng, Xuehai He, Kuan Wang, Jianfeng Gao, Weizhu Chen, Shuohang Wang, Simon~Shaolei Du, and Yelong Shen.
\newblock Reinforcement learning for reasoning in large language models with one training example, 2025.
\newblock URL \url{https://arxiv.org/abs/2504.20571}.

\bibitem[Wei et~al.(2022)Wei, Wang, Schuurmans, Bosma, Ichter, Xia, Chi, Le, and Zhou]{wei2022chain}
Jason Wei, Xuezhi Wang, Dale Schuurmans, Maarten Bosma, Brian Ichter, Fei Xia, Ed~H. Chi, Quoc~V. Le, and Denny Zhou.
\newblock Chain-of-thought prompting elicits reasoning in large language models.
\newblock \emph{arXiv preprint arXiv:2201.11903}, 2022.
\newblock URL \url{https://arxiv.org/abs/2201.11903}.

\bibitem[Yang et~al.(2025)Yang, Li, Yang, Zhang, Hui, Zheng, Yu, Gao, Huang, Lv, Zheng, Liu, Zhou, Huang, Hu, Ge, Wei, Lin, Tang, Yang, Tu, Zhang, Yang, Yang, Zhou, Zhou, Lin, Dang, Bao, Yang, Yu, Deng, Li, Xue, Li, Zhang, Wang, Zhu, Men, Gao, Liu, Luo, Li, Tang, Yin, Ren, Wang, Zhang, Ren, Fan, Su, Zhang, Zhang, Wan, Liu, Wang, Cui, Zhang, Zhou, and Qiu]{yang2025qwen3technicalreport}
An~Yang, Anfeng Li, Baosong Yang, Beichen Zhang, Binyuan Hui, Bo~Zheng, Bowen Yu, Chang Gao, Chengen Huang, Chenxu Lv, Chujie Zheng, Dayiheng Liu, Fan Zhou, Fei Huang, Feng Hu, Hao Ge, Haoran Wei, Huan Lin, Jialong Tang, Jian Yang, Jianhong Tu, Jianwei Zhang, Jianxin Yang, Jiaxi Yang, Jing Zhou, Jingren Zhou, Junyang Lin, Kai Dang, Keqin Bao, Kexin Yang, Le~Yu, Lianghao Deng, Mei Li, Mingfeng Xue, Mingze Li, Pei Zhang, Peng Wang, Qin Zhu, Rui Men, Ruize Gao, Shixuan Liu, Shuang Luo, Tianhao Li, Tianyi Tang, Wenbiao Yin, Xingzhang Ren, Xinyu Wang, Xinyu Zhang, Xuancheng Ren, Yang Fan, Yang Su, Yichang Zhang, Yinger Zhang, Yu~Wan, Yuqiong Liu, Zekun Wang, Zeyu Cui, Zhenru Zhang, Zhipeng Zhou, and Zihan Qiu.
\newblock Qwen3 technical report, 2025.
\newblock URL \url{https://arxiv.org/abs/2505.09388}.

\bibitem[Yu et~al.(2025)Yu, Zhang, Zhu, Yuan, Zuo, Yue, Fan, Liu, Liu, Liu, et~al.]{yu2025dapo}
Qiying Yu, Zheng Zhang, Ruofei Zhu, Yufeng Yuan, Xiaochen Zuo, Yu~Yue, Tiantian Fan, Gaohong Liu, Lingjun Liu, Xin Liu, et~al.
\newblock Dapo: An open-source llm reinforcement learning system at scale.
\newblock \emph{arXiv preprint arXiv:2503.14476}, 2025.

\bibitem[Zeng et~al.(2025)Zeng, Huang, Liu, Liu, He, Ma, and He]{zeng2025simplerl}
Weihao Zeng, Yuzhen Huang, Qian Liu, Wei Liu, Keqing He, Zejun Ma, and Junxian He.
\newblock Simplerl-zoo: Investigating and taming zero reinforcement learning for open base models in the wild.
\newblock \emph{arXiv preprint arXiv:2503.18892}, 2025.

\bibitem[Zhao et~al.(2022)Zhao, Li, Li, and Zhang]{zhao2022multihiertt}
Yilun Zhao, Yunxiang Li, Chenying Li, and Rui Zhang.
\newblock Multihiertt: Numerical reasoning over multi hierarchical tabular and textual data.
\newblock \emph{arXiv preprint arXiv:2206.01347}, 2022.

\end{thebibliography}
\bibliographystyle{iclr2026_conference.bst}


\end{document}